\documentclass[letterpaper, 10 pt, conference]{ieeeconf}  

\IEEEoverridecommandlockouts                              

\usepackage[utf8]{inputenc}

\usepackage{graphics} 
\usepackage{epsfig} 
\usepackage{mathptmx} 
\usepackage{times} 
\usepackage{amsmath} 
\usepackage{amssymb}  
\usepackage[dvipsnames]{xcolor}  
\usepackage[ruled, vlined, linesnumbered]{algorithm2e}  
\usepackage[font=normalsize]{caption}  
\usepackage[font=small]{subcaption}  
\usepackage{hyperref}  
\usepackage{todonotes}
\usepackage{import}
\usepackage{cite}
\usepackage{mathtools}
\usepackage{tabu}
\usepackage{dblfloatfix}
\usepackage{eso-pic}

\title{\LARGE \bf
MAPS-X: Explainable Multi-Robot Motion Planning via Segmentation
}

\author{Justin Kottinger$^{1}$, Shaull Almagor$^{2}$, and Morteza Lahijanian$^{1,3}$
\thanks{J. Kottinger received funding from the Department of Education through a Graduate Assistantships in Areas of National Need Fellowship in Critical Aerospace Technologies. S. Almagor has received funding from the European Union's Horizon 2020 research and innovation programme under the Marie Sk{\l}odowska-Curie grant agreement No 837327.}
\thanks{$^{1}$ Department of Aerospace Engineering Sciences, University of Colorado Boulder, USA
        {\tt\small \{firstname.lastname\}@colorado.edu}}%
\thanks{$^{2}$ The Henry and Marilyn Taub Faculty of
	Computer Science, Technion, Israel
        {\tt\small shaull@cs.technion.ac.il}}%
\thanks{$^{3}$ Department of Computer Science, University of Colorado Boulder, USA
        }%
}

\newcommand{\V}{V} 
\newcommand{\E}{E} 
\newcommand{\x}[1]{\mathbf{x}_{#1}}
\newcommand{\cntrl}[1]{\mathbf{u}_{#1}}
\newcommand{\U}{\mathbb{U}}
\newcommand{\X}{\mathbb{X}}

\newcommand{\R}[1]{\mathbb{R}^{#1}}
\newcommand{\dt}{\Delta T}
\newcommand{\Goal}{\mathbb{G}}
\newcommand{\xDot}[1]{\mathbf{\dot{x}}_{#1}}
\newcommand{\traj}[2]{\x{#1}^{#2}}
\newcommand{\proj}[2]{\textsc{proj}_{#1}^{#2}}
\newcommand{\mmp}{\textsc{MMP}\xspace}
\newcommand{\plannerX}{X\xspace}

\newtheorem{problem}{Problem}
\newtheorem{theorem}{Theorem}

\begin{document}
\AddToShipoutPictureBG*{%
  \AtPageUpperLeft{%
    \hspace{14.75cm}%
    \raisebox{-1.5cm}{%
      \makebox[0pt][r]{To appear in International Conference on Robotics and Automation (ICRA), May 2021.}}}}

\maketitle
\thispagestyle{empty}
\pagestyle{empty}

\begin{abstract}
%
%
Traditional \textit{multi-robot motion planning} (MMP) focuses on computing trajectories for multiple robots acting in an environment, such that the robots do not collide when the trajectories are taken simultaneously. In \emph{safety-critical} applications, a human supervisor may want to verify that the plan is indeed collision-free. 
In this work, we propose a notion of explanation for a plan of \mmp, based on visualization of the plan as a short sequence of images representing time segments, where in each time segment the trajectories of the agents are disjoint, clearly illustrating the safety of the plan. 
We show that standard notions of optimality (e.g., makespan) may create conflict with short explanations. 
Thus, we propose meta-algorithms, namely \textit{multi-agent plan segmenting}-\plannerX (MAPS-\plannerX) and its lazy variant, that can be plugged on existing centralized sampling-based tree planners \plannerX to produce plans with good explanations using a desirable number of images. We demonstrate the efficacy of this explanation-planning scheme and extensively evaluate the performance of MAPS-\plannerX  
and its lazy variant in various environments and agent dynamics.  
\end{abstract}

\section{Introduction}
  \label{sec:intro}

\textit{Multi-robot motion planning} (\mmp) is a fundamental challenge in robotics and \textit{artificial intelligence} (AI).  The goal in \mmp is to plan trajectories for multiple robots according to their dynamics to reach their respective goal regions such that, when the plans are executed simultaneously, every robot (agent) successfully completes its trajectory without colliding with other agents or obstacles. Applications of \mmp can be found in many areas where several moving agents interact in a shared workspace. 
One limitation of various AI tools, including \mmp, is their 
inability to explain their decisions and actions to human users \cite{turek}. In many \emph{safety-critical} application domains, such as air-traffic control and hazardous material warehouses, \mmp tools are rarely utilized if at all.
Instead, the trajectories are either hand designed or, if a planner is used, the generated paths are given to a human-supervisor for safety verification before execution.  
Thus, these settings require the plan to be presented in a humanly-understandable manner. Specifically, the presentation should enable the supervisor to understand the path taken by individual agents and to verify that the agents do not collide. To this end, the goal of this work is to present a method of generating explainable motion plans for multi-agent systems.

In general, \mmp is \textbf{NP-Complete} even in the discrete (finite space) setting and naturally becomes intractable in the continuous (space) setting as the number of robots grows. A significant body of work is dedicated to overcoming this difficulty, both in the discrete domain (e.g., \cite{stern2019mapf,standley2010finding}) and continuous domain (e.g., \cite{gravot2003method,gharbi2009roadmap,wagner2015subdimensional,shome2020drrt}).  
There are two general approaches to \mmp: \textit{centralized} methods, which work in the composed space of all the agents \cite{solovey2015finding,wagner2015subdimensional,shome2020drrt}, and \textit{de-centralized} approaches, which divide the problem into several subproblems and solve each separately \cite{gammell2014bit,sanchez2002using,van2005prioritized,tang2018complete}.
The focus of all these works is to overcome the general difficulty of computing a plan in a short amount of time and optimizing the plan according to a measure such as makespan and path length. 
These works do not take into consideration the explainability of the plan. In fact, explainability often conflicts with other optimality measures and requires separate treatment.

\begin{figure}[t]
	\centering
	\begin{subfigure}{.48\linewidth}
		\centering
		\includegraphics[width=\textwidth]{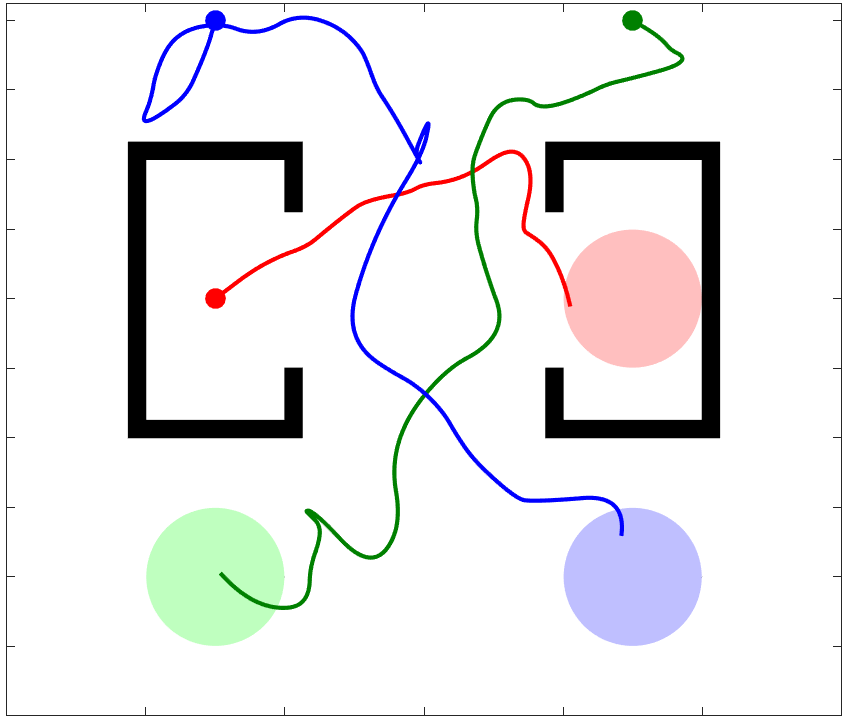}
		\caption{Full Plan}
		\label{fig:FullPlan}
	\end{subfigure}
	~
	\begin{subfigure}{.48\linewidth}
		\centering
		\includegraphics[width=\textwidth]{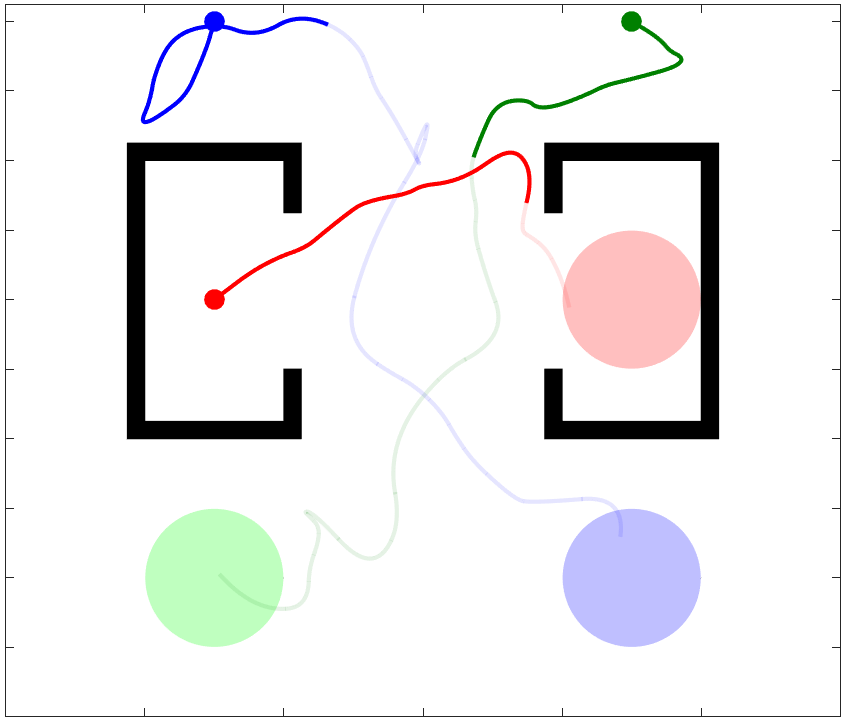}
		\caption{Part 1}
		\label{fig:part1}
	\end{subfigure}

	\vspace{1mm}
	
	\begin{subfigure}{.48\linewidth}
		\centering
		\includegraphics[width=\textwidth]{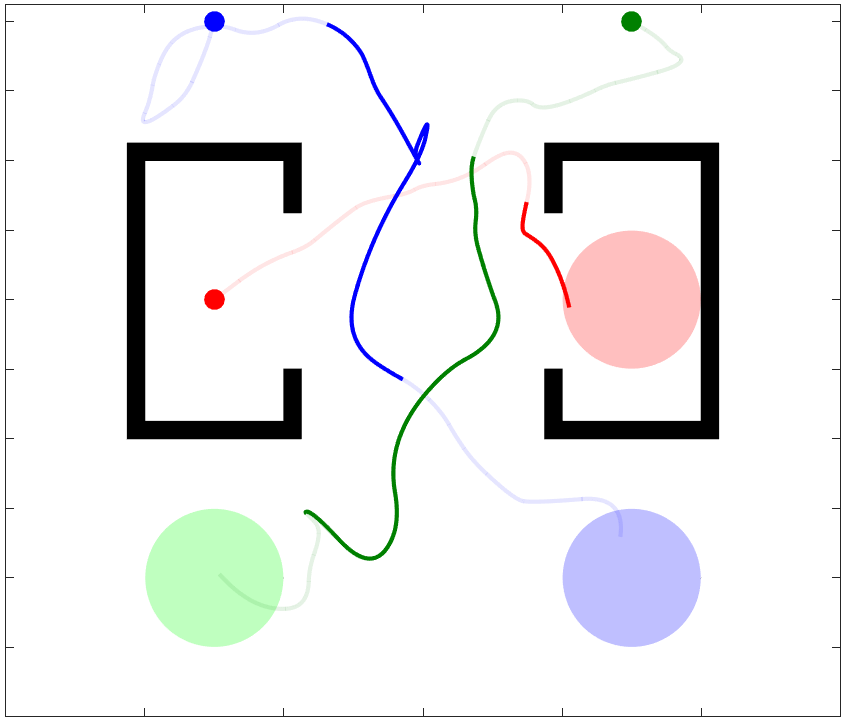}
		\caption{Part 2}
		\label{fig:part2}
	\end{subfigure}
	~
	\begin{subfigure}{.48\linewidth}
		\centering
		\includegraphics[width=\textwidth]{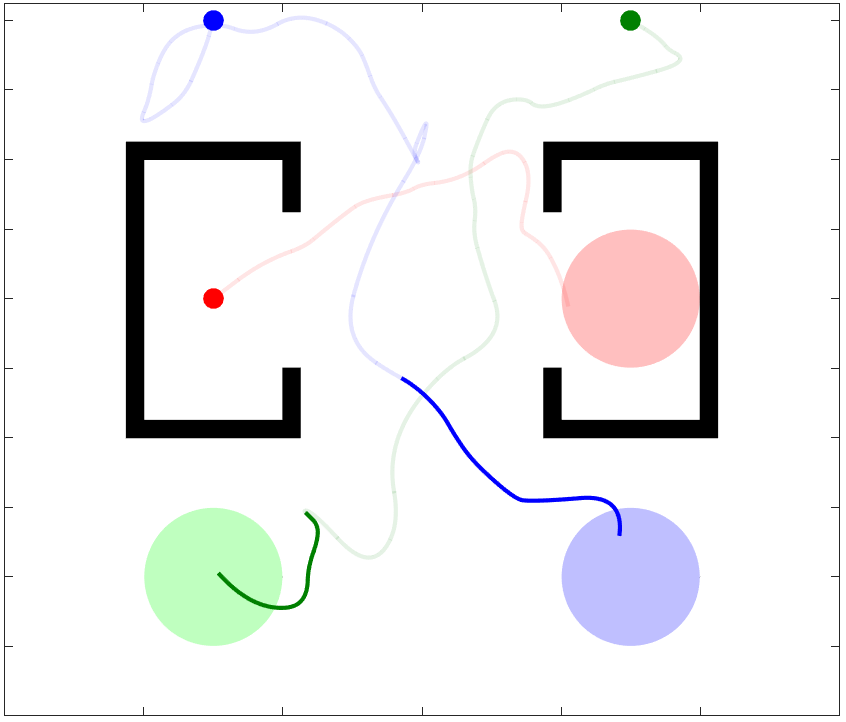}
		\caption{Part 3}
		\label{fig:part3}
	\end{subfigure}
	\caption{\small A plan for three (second-order) robotic agents in (a) is explained for visual verification via disjoint decomposition in (b)-(d). The small and large circles mark the initial and goal locations for the agents, respectively. 
	}
	\label{fig:explain-graph}
	\vspace{-5mm}
\end{figure}

Significant effort has been dedicated to providing explanations for problems in AI and machine learning. 
For example, work \cite{Lapuschkin_2019} utilizes visualization to explain the result of certain machine learning algorithms that often come up with complicated classifiers. In \cite{eifler_cashmore_hoffmann}, explanations are given by analyzing alternative plans with some user-defined properties. In \cite{10.5555/3306127.3331663}, a user proposes a plan, and explanations are given as a minimal set of differences between the actual plan and the proposed plan. A broader approach was later given in \cite{fox2017explainable}, where multiple types of explanations are allowed.
In that setting, the user can change the plan, motivating the planner to either explain why the original plan is better or to re-plan.  None of these studies, however, focus on the \mmp problem.

In~\cite{almagor2020explainable}, we propose an explanation scheme based on visualization for \mmp in discrete domains. There, in order to convince a human supervisor that a suggested plan does not cause a collision between the agents, the plan is decomposed into time segments, such that within each segment the paths of the agents are \emph{disjoint} (i.e., non-intersecting). Then, an explanation of the plan comprises a sequence of images representing each segment.  An example of such explanations for three continuous agents is shown Fig.~\ref{fig:explain-graph}.
It is important to note that, since identification of line intersections is made very early in the cognitive process (namely in the primary visual cortex)~\cite{Hubel,Tang}, it is easy for a human to verify that in each segment, the paths do not intersect. Moreover, the sequence of images is potentially (and indeed, often in practice) much shorter than displaying, e.g., a slowed-down video of the agents taking their paths, and is hence easy to verify.
In addition,  \cite{almagor2020explainable} showed that finding optimal explanations for a given motion plan can be done in polynomial time, whereas generating plans for explainability, i.e., limiting the number of segments, is, at best, \textbf{NP-Complete}.

The algorithms proposed in~\cite{almagor2020explainable} are based on a discrete (or discretized) environment, and thus overlook the challenges involved with motion planning in the continuous domain.
In this work, we focus on \mmp with explanations for realistic robotic systems in the continuous space with kinodynamical constraints. 
To this end, we treat explainability as an additional constraint on top of \mmp, and incorporate it into existing sampling-based algorithms.  As mentioned above and shown in~\cite{almagor2020explainable}, there is often a trade-off between planning for short explanations and short paths. Hence, explainability may conflict with state-of-the-art heuristics for \mmp.  In order to factor out precise heuristics, we devise generic meta-algorithms, that search for optimally-explainable plans using any centralized sampling-based algorithm.  
We demonstrate our meta algorithms by plugging them with classical motion planners such as \emph{rapidly-exploring random trees} (RRT) \cite{Lavalle98rapidly-exploringrandom}.

The main contribution of this work is an explanation scheme for \mmp that is based on path segmentation in the continuous domain.  This scheme
introduces a new constraint (challenge) to the motion planning problem that is not previously studied, to the best of our knowledge.  We present two meta algorithms called \textit{multi-agent plan segmenting}-\plannerX (MAPS-\plannerX), where \plannerX can be any existing centralized (kinodynamic) \mmp planner.  Another contribution is an extensive evaluation of these algorithms, highlighting their generality and differences in performance.

\section{Problem Formulation}
  \label{sec:problem}

Consider $k \in \mathbb{N}$ robotic systems (agents), in a shared workspace $W\subseteq \R{2}$ which includes a finite set of obstacles $O$, where each obstacle $o \in O$ is a closed subset of $W$, i.e., $o\subset W$. 
The motion of each agent $i \in \{1, 2, \ldots , k\}$ is subject to the following dynamic constraint:
\begin{align}
  \xDot{i} & = f_i (\x{i}, \cntrl{i}),           &  \x{i}&\in X_i \subseteq \R{n_i},              &  \cntrl{i}&=U_i \subseteq \R{m_i},
  \label{eq:sys}
\end{align}

\noindent
where $X_i$ and $U_i$ are the agent $i$'s state and input spaces, respectively, and $f_i:X_i \times U_i \rightarrow X_i$ is an integrable and possibly nonlinear function.

Given a time interval
$[t_0, t_f]$, where $t_0,t_f \in \R{}_{\geq 0}$ and $t_0 < t_f$, 
a controller $\cntrl{i}: [t_0,t_f] \to U_i$,
and initial state $x_{i,0} \in X_i$, 
function $f_i$ can be integrated up to time $t_1 \leq t_f$ to form a \textit{trajectory segment} $\traj{i}{t_0: t_1}$ for agent $i$, where $\traj{i}{t_0: t_1}(t_0)=x_{i,0}$. 
For $s\in \mathbb{N}$ consecutive time intervals,
\begin{equation}
  \label{eq:time-intervals}
  [t_0, t_1],[t_1,t_2], \ldots, [t_{s-1}, t_s],
\end{equation}
we define a \textit{trajectory} 
$$T_i=\{\traj{i}{t_0: t_1}, \traj{i}{t_1: t_2}, \ldots, \traj{i}{t_{s-1}: t_s}\},$$ 
where 
$\traj{i}{t_{l-1}: t_{l}}(t_l) = \traj{i}{t_{l}: t_{l+1}}(t_{l})$ for all $l \in \{1,2,\ldots,s-1\},$
to be a set of $s$ trajectory segments.  


Let $X_i^G \subset X_i$ and $x_{i, 0} \in X_i$ denote the goal region (destination) and initial state of  agent $i$, respectively.
The goal multi-robot motion planning (\mmp) is to find a trajectory $T_i$ with $\x{i}(t_0) = x_{i, 0}$ for every agent $i \in \{1, 2, \dots, k\}$ such that no agent collides with any obstacles nor with other agents, and $\x{i}(t_f) \in X_i^G$.  
In \emph{explainable} \mmp we require that, in addition, the interval $[t_0,t_f]$ can be segmented to at most $r\in \mathbb{N}$ consecutive time intervals (for some user-defined upper bound $r$), such that within each time interval, the trajectories are disjoint when presented to the user.

To formally define the explainable \mmp problem, we start by formalizing the notion of disjoint segments.
Let 
$\proj{W}{X_i}: \R{n_i}\rightarrow W$ be a function that projects a 
state $x_i \in X_i$ of agent $i$
onto workspace $W$.  Then, we call two trajectory segments $\traj{i}{t_1:t_2}$ and $\traj{j}{t_1:t_2}$ \textit{disjoint} if 
$$\proj{W}{X_i}(\traj{i}{t_1:t_2}(t)) \; \neq \; \proj{W}{X_j}(\traj{j}{t_1:t_2}(t')) \quad \forall t,t' \in [t_1,t_2].$$


We extend the notion of disjoint from segments to trajectories as follows.  
Given the $s$ time intervals from \eqref{eq:time-intervals}, 
we call a set of trajectories $T = \{T_1,T_2, \ldots, T_k\}$ \emph{segment-disjoint} 
if for every segment $l \in \{ 1, 2, \ldots, s\}$ the induced trajectory-segments $\traj{i}{t_{l-1}:t_{l}} \in T_i$ and $\traj{j}{t_{l-1}:t_{l}} \in T_j$ are disjoint for all  $i,j \in \{1, 2, \ldots,k \}$ and $i \neq j$. 

We can now formulate the \emph{explainable} \mmp problem as follows.

\begin{problem}[Explainable \mmp]
  \label{problem}
  Given $k$ robotic agents with dynamics described in \eqref{eq:sys}, initial states 
  $x_{1,0},\ldots,x_{k,0}$, 
  goal regions $X^G_1, \ldots, X^G_k$, and a bound $r \in \mathbb{N}$ on the number of segments, find a controller $\cntrl{i}:[t_0,t_f] \to U_i$ for each agent $i\in\{1,\ldots,k\}$ and time points $t_1,\ldots,t_s$, where $s \leq r$ and $t_0 < t_1 < \ldots < t_{s-1} < t_s = t_f$, such that the obtained trajectory $T_i$ takes agent $i$ from $\x{i}(t_0) = x_{i,0}$ to $\x{i}(t_f) \in X^G_i$ while avoiding collisions with obstacles, and the set of trajectories $T =\{T_1, \ldots, T_k\}$ is segment-disjoint. 
\end{problem}

Note that the segment-disjoint requirement for the trajectories in Problem \ref{problem} implies that the trajectories do not cause collisions between the agents.  Further, a solution to this problem consists of not only $k$ valid trajectories but also a decomposition of these trajectories into $s$ set of disjoint segments. 
The explanations are then $s$ images, one image per set of segments projected onto the workspace $W$, e.g., Fig. \ref{fig:part1}-\ref{fig:part3}.


We emphasize that our goal in this work is not to design an efficient algorithm that solves the general \mmp problem.  Rather, our goal is to design meta algorithms that turn an existing \mmp planner into an explainable \mmp that solves Problem~\ref{problem}.
Specifically, we focus on centralized \mmp, where planning takes place in
the compound space of the agents. 
This approach has two advantages: first, centralized algorithms maintain most of the structure of the search space. This allows us to obtain optimal explanations (segments), which in turn sets a baseline by which to compare more involved algorithms for explainable \mmp.  Second, centralized planning uses well-understood algorithms, which enable us to reason about the explanations we obtain with respect to different algorithms. 
In contrast, de-centralized \mmp algorithms may be too involved to separate their properties from the explanations they yield. We discuss incorporating explanations into more involved algorithms in Section~\ref{sec:conclusion}.

Finally, we remark that our definitions of disjoint trajectory segments is based on trajectory intersections.  This definition (and the proposed algorithms) can be easily extended to intersections of robots with 2D shapes.



\section{Algorithms}
  \label{sec:algorithms}

In this section, we present three methodologies to solve Problem~\ref{problem}. 
These are based on a centralized approach to \mmp, so we first present Planner \plannerX, which is a generic centralized sampling-based tree planner.
Then, we define a post-process procedure that can minimally decompose (i.e., explain) an existing \mmp solution, e.g., returned by Planner \plannerX, into disjoint segments. 
Next, we outline two frameworks that can be incorporated into 
Planner \plannerX
to solve explainable \mmp queries.  


\subsection{Planner \plannerX: Centralized Sampling-based Tree \mmp}
\label{sec:Xplanner}
Centralized tree-based planners grow a motion tree in the composed state space $\X = X_1 \times X_2 \times \ldots \times X_k$ of the agents according to the dynamics in \eqref{eq:sys}
through sampling and extension procedures.
A generalized sampling-based tree planner \plannerX is outlined in Alg.~\ref{alg:X}. It takes the composed state and input spaces, $\X$ and $\U = U_1 \times \ldots \times U_k$, respectively, the goal set $\Goal = X_1^G \times \ldots \times X_k^G$, the set of obstacles $O$, an initial configuration for all agents $x_{0}  = (x_{1, 0}^T,\ldots,x_{k, 0}^T)^T \in \X$, and a specified number of iterations $N \in \mathbb{N}$ and returns a valid trajectory if found in $N$ iterations.

\begin{algorithm}[h]
 $G = \{\V \leftarrow x_{0}, \E \leftarrow \emptyset \}$\;
 \For{$N$ \textit{iterations}}{
  $(x_{rand},x_{near}) \leftarrow sample(\X, \V)$\; \label{line:sampleNode}
  $x_{new} = \textsc{multiAgentExtend}(x_{near}, x_{rand}, \mathbb{U})$\; \label{line:maInt}
  \If{$isValid(\overrightarrow{x_{near}, x_{new}})$}{
   $\V \leftarrow \V \cup \{x_{new}\};$ $\E \leftarrow \E \cup \{\overrightarrow{x_{near}, x_{new}}\}$\; \label{line:add2G}
   \If{$x_{new}\in \Goal$}
   {
   \KwRet{$\overrightarrow{x_{0}, x_{new}}$}\;
   }
   }
 }
 \KwRet{$G(\V, \E)$}
 \caption{Planner \plannerX ($\X$, $\U$, $\Goal$, O, $x_{0}$, N)}
 \label{alg:X}
\end{algorithm}

The algorithm first initializes the graph with the initial state (root node) $x_{0}$. Next, 
an existing node $x_{near}$ and a random state $x_{rand}$ are picked through a sampling process in Line~\ref{line:sampleNode}.
After that, Line~\ref{line:maInt} samples control inputs $u \in \U$ and propagates the multi-agent system from $x_{near}$ toward $x_{rand}$ to generate a new state $x_{new}$. 
If the newly generated trajectory $\overrightarrow{x_{near}, x_{new}}$ is valid, i.e., it does not result in a collision with obstacles nor with other agents, then Line~\ref{line:add2G} adds the vertex $x_{new}$ and the edge $\overrightarrow{x_{near}, x_{new}}$ to $G$. This process repeats a maximum of $N$ times, exiting early if a solution is found. To speed up computation, it is common in the multi-agent setting to  propagate only agents $i$ that are \emph{not} in their respective goal region $X_i^G$ in the procedure $\textsc{multiAgentExtend}$.
For the remainder of the paper, we refer to this planner as Planner \plannerX.

\subsection{Minimal Disjoint Segmentation} 
\label{sec:segmentation}
One method of explaining a multi-agent trajectory is to solve the explainability problem \emph{after} the planning problem. This solution
involves running a motion planner (e.g., Planner \plannerX) to generate a solution $T = \overrightarrow{x_{0}, x_{goal}}$, and then finding a minimal segmentation of the solution. This post-processing procedure, shown in Alg. \ref{alg:postProcess}, is called \emph{\textsc{segmentSol}}. 

\begin{algorithm}[htb]{}
	\SetKwInOut{Input}{input}\SetKwInOut{Output}{output}
	$d$, $s \leftarrow 1$;  \label{line:setParams}
	$v \leftarrow x_{0}$\;
	\While{$v \neq x_{goal}$}
	{
		$intersection \leftarrow \textsc{project2D}(v, d)$\; \label{line:proj2D}
		\eIf{intersection}
		{
			\For{$d - 1$ $times$}
			{
				$v.segmentNum \leftarrow s$;\label{line:makeSegment}  $v \leftarrow v.child$\;\label{line:endMakeSegment}
			}
			$d \leftarrow 1$; $s \leftarrow s + 1$\; \label{line:incSegment}
		}
		{
			$d \leftarrow d + 1$\; \label{line:addDepth}
		}
	}
	\KwRet{$\overrightarrow{x_{0}, x_{goal}}$}
	\caption{\textsc{segmentSol}($T = \protect\overrightarrow{x_{0}, x_{goal}}$)}
	\label{alg:postProcess}
\end{algorithm}
\vspace{-3mm}

Alg. \ref{alg:postProcess} is implemented using a greedy approach -- we traverse the nodes of the solution trajectory $\overrightarrow{x_{0}, x_{goal}}$, and add nodes to a segment as long as the projection of the agents' trajectories do not intersect (Line~\ref{line:proj2D}). 
Once an intersection occurs, we end the segment (collating the nodes before the intersection), and start a fresh segment (Lines~\ref{line:makeSegment} and \ref{line:incSegment}). 

Checking for intersections (procedure \textsc{project2D}, Line~\ref{line:proj2D}) can be implemented efficiently, e.g., using linear interpolations of the projected paths in some $\Delta t$ intervals.
As proven in~\cite{almagor2020explainable}, the greedy approach of Alg. \ref{alg:postProcess} is guaranteed to obtain a minimal segmentation among those whose end-points are multiples of $\Delta t$. Thus, we have the following theorem.
\begin{theorem}[Min. disjoint segmentation \cite{almagor2020explainable}, Thm. 3.1]  
	\label{th:minSegmentation}
	Given a set of trajectories $T = \{T_1,\ldots,T_k\}$ for $k$ agents, Alg. \ref{alg:postProcess} computes a segmentation of $T$ such that the resulting trajectories are segment-disjoint with minimal number of segments\footnote{Minimal among segmentations whose end-points are multiples of $\Delta t$.}.
\end{theorem}

Observe that Alg. \ref{alg:postProcess} makes a single pass on the trajectory, but for each new node along the trajectory, intersections need to be checked against the segment collated so far. Thus, the algorithm has a quadratic running time, parametrized by the implementation of $\textsc{project2D}$, 
which can be made efficient.
The post-processing procedure \textsc{segmentSol} provides explanations for a solution of the \mmp problem, i.e., enables users to validate that multi-agent trajectories are collision free.
While this procedure guarantees a minimal segmentation for the given trajectory, it cannot guarantee that this number 
is below the given bound $r$ (or indeed, the minimal segmentation of any trajectory). That is because the planning process and the segmentation process are completely separate. Thus, strictly relying on Alg.~\ref{alg:postProcess} could result in unsatisfiable explanations that require many segments. 
Next, we show how to solve the planning problem and the explanability problem simultaneously. 

\subsection{Planning with Segmentation}

\subsubsection{Lazy MAPS-\plannerX}
\label{sec:lazy-maps}
The most intuitive solution for combining explanations into the planning procedure is to incorporate Alg.~\ref{alg:postProcess} into Planner \plannerX.
The resulting algorithm is shown in Alg.~\ref{alg:lazyalg:MapsX}. It runs Planner \plannerX until a solution is found. Then, rather than immediately exiting the loop with a solution, Line~\ref{line:LazyXsegment} calculates the number of disjoint segments of the solution. 
If it is satisfactory, i.e., it is at most the user-defined upper bound $r$,  it returns trajectory $T = \overrightarrow{x_{0}, x_{new}}$. Otherwise, Line~\ref{line:LazyXprune} prunes the unsatisifiable portion of the solution and continues planning. The loop repeats until a satisfiable solution is found. 
\vspace{-3mm}

\begin{algorithm}[htb]
\SetKwInOut{Input}{input}\SetKwInOut{Output}{output}
 $G = \{\V \leftarrow x_{0}, \E \leftarrow \emptyset \}$\;
 \For{$N$ \textit{iterations}}{
  $(x_{rand},x_{near}) \leftarrow sample(\X, \V)$\; \label{line:LazyXsampleNode}
  $x_{new} = \textsc{multiAgentExtend}(x_{near}, x_{rand}, \mathbb{U})$\; \label{line:LazyXmaInt}
  \If{$isValid(\overrightarrow{x_{near}, x_{new}})$}{
   $\V \leftarrow \V \cup \{x_{new}\}$; $\E \leftarrow \E \cup \{\overrightarrow{x_{near}, x_{new}}\}$\;
   \If{$\x{new}\in \Goal$}
   {
   $s \leftarrow \textsc{segmentSol}(\overrightarrow{x_{0}, x_{new}})$\; \label{line:LazyXsegment}
   \eIf{$s \leq r$}
   {
    \KwRet{$\overrightarrow{x_{0}, x_{new}}$}
   }
   {
    $pruneSol(\overrightarrow{x_{0}, x_{new}})$\; \label{line:LazyXprune}
   }
   }
   }
 }
 \KwRet{$G(\V, \E)$}
 \caption{Lazy MAPS-\plannerX($\X$, $\U$, $\Goal$, O, $x_{0}$, N, $r$)}
 \label{alg:lazyalg:MapsX}
\end{algorithm}
\vspace{-3mm}

We call this framework \textit{lazy multi-agent plan segmenting} \plannerX (Lazy MAPS-\plannerX) because it `lazily' turns Planner \plannerX into an explainable planner. 
The number of segments required to explain a trajectory segment from the root node $x_{0}$ to a node $x_{new}$ is \emph{only} considered if it is part of a possible solution.  All other nodes are ignored. 
In  Section~\ref{sec:experiments}, we show  that Lazy MAPS-\plannerX works well when the environment naturally admits small number of explanations, but its benefits are hindered when the system behavior must be changed to match optimal explanation schemes. In such a situation, a more involved framework is desired.

\subsubsection{MAPS-\plannerX}
\label{sec:maps}
We provide another version of our framework by further integrating segmentation into the planner.
This planner differs from its Lazy counterpart by calculating how many segments are required to explain each trajectory from the root node $x_0$ to a new node $x_{new}$ during the planning phase. In doing so, MAPS-\plannerX only adds nodes that have a satisfiable number of explanations to graph (tree) $G$. 

Alg.~\ref{alg:MapsX} outlines MAPS-\plannerX. 
Here, we define a \emph{cost} for each node $x_{new}$ in the graph $G$ equivalent to the number of segments required to explain the trajectory from the root node $x_{0}$ to $x_{new}$. After verifying that $x_{new}$ is valid, MAPS-\plannerX calculates the cost of the node in Line~\ref{line:MAPScalcCost} through procedure \textsc{findTotalCost}. If it is satisfiable ($x_{new}.cost \leq r$), then the node is added to the tree; otherwise, it is rejected. 

Procedure \textsc{findTotalCost} computes the cost of a new node by calling Alg.~\ref{alg:postProcess}.
For every new node $x_{new}$, \textsc{findTotalCost} first identifies the current trajectory segment, i.e., set of nodes with the same cost (segment number) as the parent of $x_{new}$ on the current branch of $G$.  Then, it calls Alg.~\ref{alg:postProcess} on the current segment to check whether the addition of $x_{new}$ requires a new segmentation, i.e., whether $\overrightarrow{x_{near},x_{new}}$ intersects with the current segment. If so, it updates the costs of the nodes on the current segment accordingly; otherwise, the cost of $x_{new}$ is set to the cost of its parent node, i.e., $x_{new}.cost = x_{new}.parent.cost$.  Note that this procedure is efficient since it only checks for the current segment, not the entire trajectory from the root to $x_{new}$.
\vspace{-3mm}

\begin{algorithm}[htb]
\SetKwInOut{Input}{input}\SetKwInOut{Output}{output}
 $G = \{\V \leftarrow x_{0}, \E \leftarrow \emptyset \}$\;
 \For{$N$ \textit{iterations}}
 {
  $(x_{rand},x_{near}) \leftarrow sample(\X, \V)$\; 
  $x_{new} = \textsc{multiAgentExtend}(x_{near}, x_{rand}, \mathbb{U})$\;
  \If{$isValid(\overrightarrow{x_{near}, x_{new}})$}
  {
    $\textsc{findTotalCost}(x_{new})$\; \label{line:MAPScalcCost}
    \If{$x_{new}.cost \leq r$}
    {
      $\V \leftarrow \V \cup \{x_{new}\}$; $\E \leftarrow \E \cup \{\overrightarrow{x_{near}, x_{new}}\}$\;
      \If{$x_{new}\in \Goal$}
      {
      \KwRet{$\overrightarrow{x_{0}, x_{new}}$}\;
      }
    }
  }
 }
 \KwRet{$G(\V, \E)$}
 \caption{MAPS-\plannerX($\X$, $\U$, $\Goal$, $x_{0}$, N, $\dt$, $r$)}
 \label{alg:MapsX}
\end{algorithm}
\vspace{-3mm}

The resulting behavior is a planner that tracks the number of segments required to explain each of its branches as the tree grows. Because only satisfiable nodes are added to $G$, MAPS-\plannerX guarantees that the number of disjoint segments (explanations) of a solution is less than or equal to $r$. 
Furthermore, since all nodes individually track their segment count (cost), once a solution is found, 
the segmentation information is already embedded in the solution, and no further computation is needed.

\subsubsection{Completeness and Optimality}

It is important to note that Alg.~\ref{alg:lazyalg:MapsX} and \ref{alg:MapsX} inherent the completeness property of the underlying Planner \plannerX. 
For example, a common property for sampling-based planners is \textit{probabilistically completeness}, 
i.e.,
as number of planning iterations $N$ approaches infinity, the probability of finding a solution, if one exists, approaches $1$. 
Then, we can make the following statement for Lazy MAPS-\plannerX and MAPS-\plannerX.
\begin{theorem}[Completeness]
  \label{th:completeness}
  Planners Lazy MAPS-\plannerX and MAPS-\plannerX are probabilistically complete if and only if Planner \plannerX is probabilistically complete.
\end{theorem}

The proof for MAPS-\plannerX follows from Theorem~\ref{th:minSegmentation} (the computed cost of each node is the minimal number of disjoint segments from root to that node) and the fact that MAPS-\plannerX mimics the behavior of Planner \plannerX, while only being more restrictive in the validity of nodes in the graph $G$ with respect to bound $r$, i.e., it adds a node to the tree only if the number of disjoint segments to the node from the root is less than or equal to $r$.  Similarly,  Lazy MAPS-\plannerX mimics the behavior of Planner \plannerX. But rather than being more restrictive in the validity of nodes in the graph $G$ during the extension procedure, Lazy MAPS-\plannerX is more restrictive in the solutions of $G$ with respect to $r$ and prunes the tree in a post procedure.
Therefore, if a solution to Problem~\ref{problem} exists in a particular query, Lazy MAPS-\plannerX and MAPS-\plannerX will find it with probability approaching $1$ as $N$ approaches infinity.

Lazy MAPS-\plannerX and MAPS-\plannerX can 
be turned into asymptotically optimal planners.
This is achieved by iteratively running a particular planning query and lowering the bound $r$ at each run. 
Hence, the resulting meta-planner monotonically decreases its cost (explanations) and hence asymptotically approaches the optimal value.

\section{Experiments and Benchmarks}
  \label{sec:experiments}

In this section, we demonstrate the efficacy of our explanation scheme and evaluate the performance of the proposed meta-algorithms Lazy MAPS-\plannerX and MAPS-\plannerX.
We implemented these algorithms with the classical motion planners 
RRT \cite{Lavalle98rapidly-exploringrandom} 
and EST \cite{hsu1997path} 
and evaluated their performance in several environments and agents dynamics. 
The benchmarking results are shown in Table~\ref{tab:table_benchmark_unlimited}.  

Our
implementations use the \textit{Open Motion Planning Library} (OMPL) \cite{sucan2012the-open-motion-planning-library} and are available here
\cite{mapsx-code}.
The benchmarks were performed on a machine with AMD Ryzen 7 3.9 GHz CUP and 64 GB of RAM.

\subsection{MAPS Planning}
We begin by showcasing our explanation scheme in several settings, and gaining some insight into its properties.

\begin{figure}[t]
     \centering
     \begin{subfigure}{0.32\linewidth}
        \includegraphics[width=\linewidth]{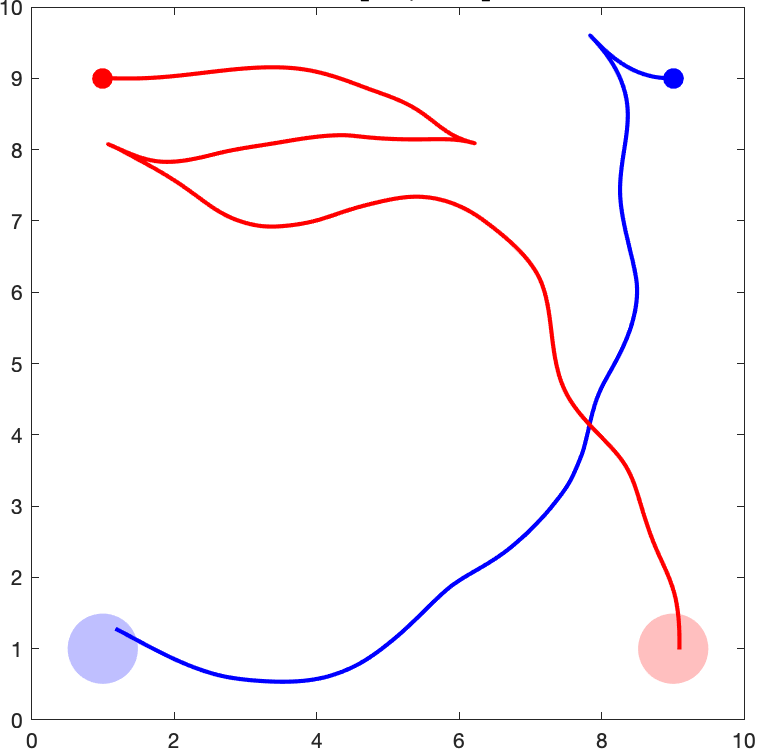}
        \caption{Full Solutions}
        \label{example1_fullPlan_2ndOrderCar}
     \end{subfigure}
     \begin{subfigure}{0.32\linewidth}
         \centering
         \includegraphics[width=\textwidth]{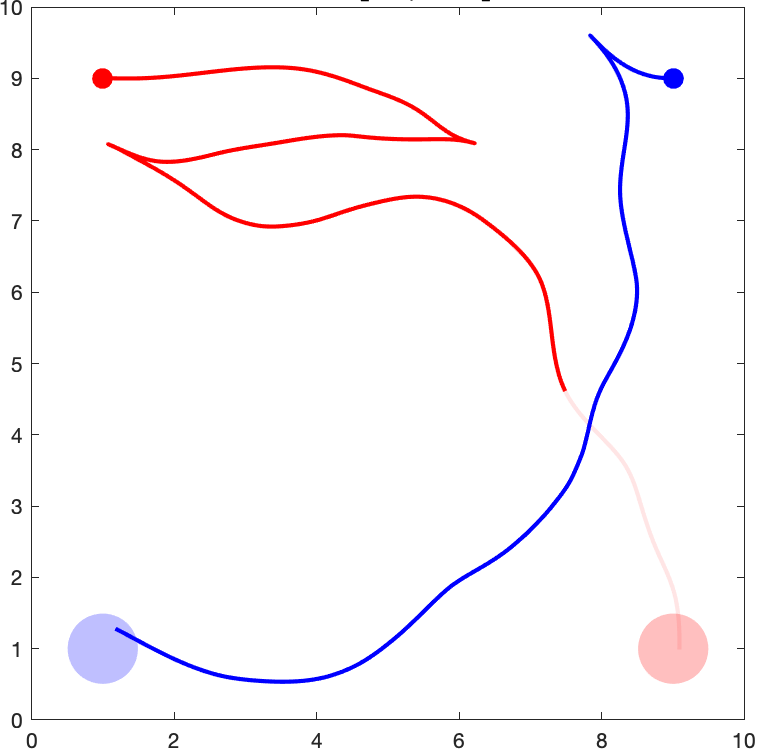}
         \caption{$\dt_1 = [0, 38.5]$}
         \label{fig:ex1seg1}
     \end{subfigure}
     \begin{subfigure}{0.32\linewidth}
         \centering
         \includegraphics[width=\textwidth]{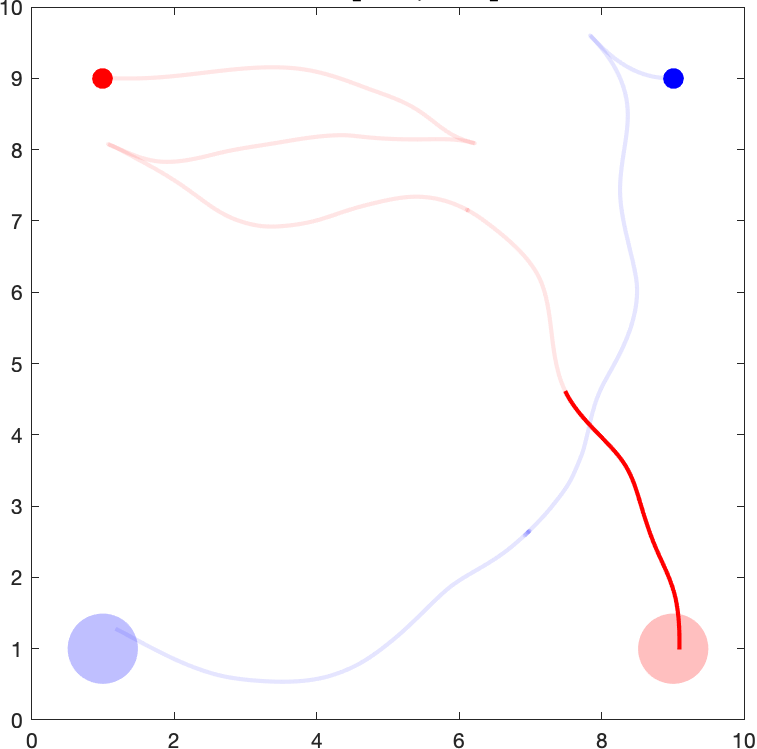}
         \caption{$\dt_2 = [38.5, 45.5]$}
         \label{fig:ex1seg2}
     \end{subfigure}
     \caption{Open Space: solutions via RRT and MAPS-RRT.}
     \label{fig:explanationEx1}
     \vspace{-5mm}
\end{figure}

Fig.~\ref{example1_fullPlan_2ndOrderCar} shows an example solution of the continuous \mmp problem produced by RRT. The example shows two agents, each with $2^\text{nd}$-order car dynamics, that cross each others' paths to get to their goals (indeed, the paths must cross in this environment).
While it is difficult for a human to validate that this plan is collision free, this becomes easy with MAPS-RRT, using two images (Figs.~\ref{fig:ex1seg1}-\ref{fig:ex1seg2}). Observe that in this example, the plan found by RRT already admits a minimal decomposition, therefore MAPS-RRT merely finds it.
\begin{figure}[t]
     \centering
     \begin{subfigure}[b]{0.32\linewidth}
         \centering
         \includegraphics[width=\linewidth]{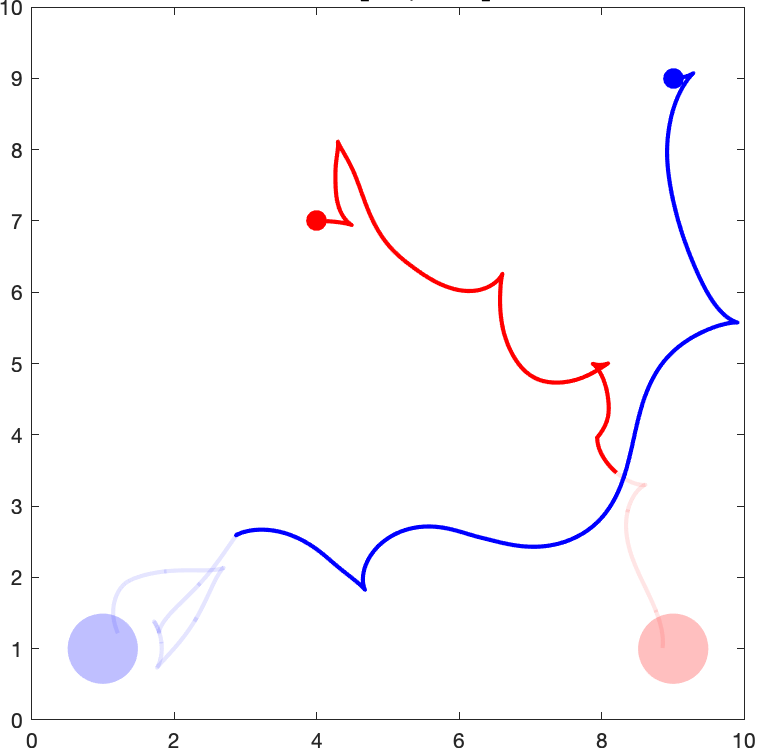}
         \caption{$\dt_1 = [0, 22.1]$}
         \label{fig:ex2Seg1}
     \end{subfigure}
     \hfill
     \begin{subfigure}[b]{0.32\linewidth}
         \centering
         \includegraphics[width=\linewidth]{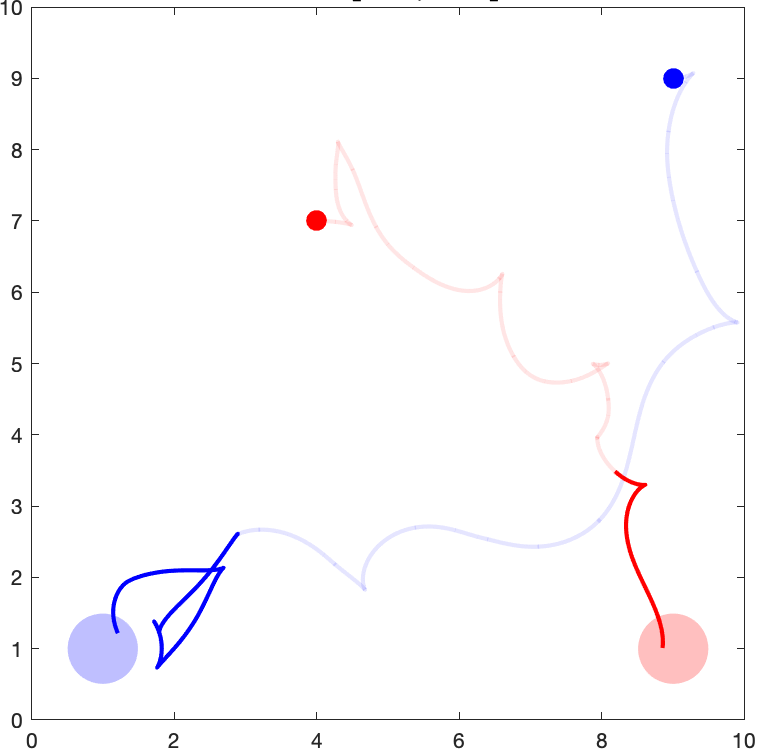}
         \caption{$\dt_2 =[22.1, 48.9]$}
         \label{fig:ex2Seg2}
     \end{subfigure}
     \begin{subfigure}[b]{0.32\linewidth}
        \centering
        \includegraphics[width=\linewidth]{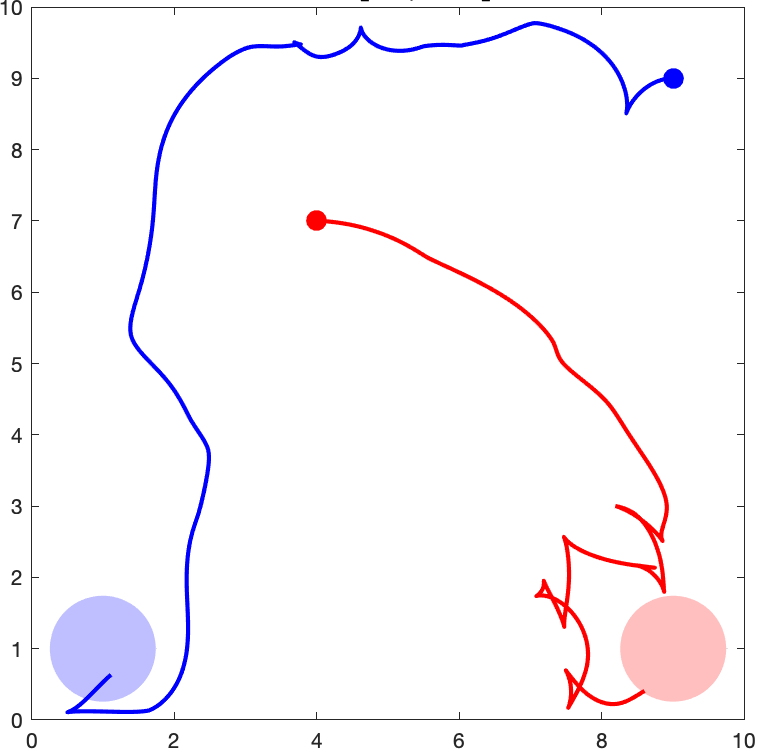}
        \caption{MAPS-X $r=1$.}
        \label{fig:ex2_OneSegment}
    \end{subfigure}
     \caption{Solution via RRT and MAPS-RRT with $r=1$.}
     \label{fig:ex2Unicylce}
     \vspace{-5mm}
\end{figure}
\begin{figure}[b]
     \centering
     \begin{subfigure}[b]{0.49\linewidth}
        \centering
        \includegraphics[width=\linewidth]{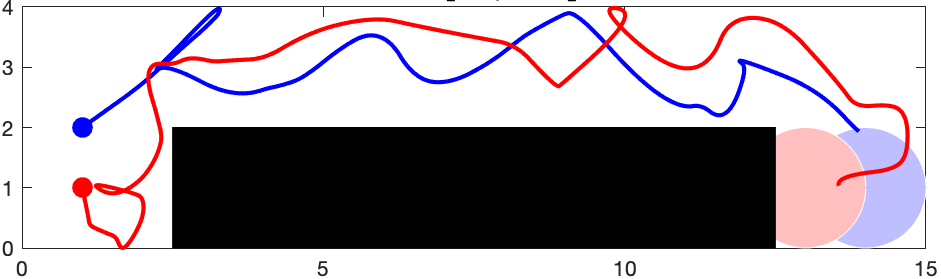}
        \caption{MAPS-RRT with $r=3$}
        \label{fig:ex4Full3}
     \end{subfigure}
     \begin{subfigure}[b]{0.49\linewidth}
        \includegraphics[width=\linewidth]{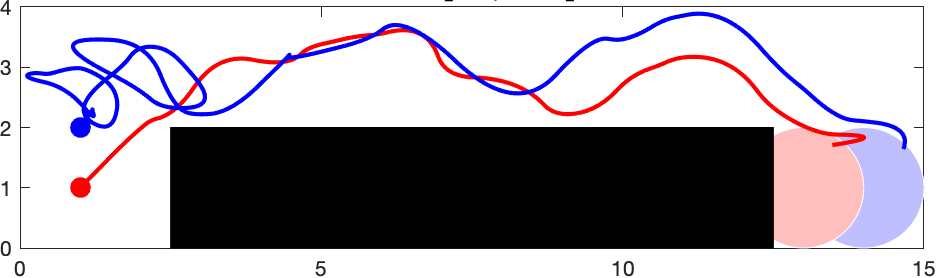}
        \caption{MAPS-RRT with $r=2$}
        \label{ex4_2ndOrderLinear}
     \end{subfigure}
     \begin{subfigure}[b]{0.49\linewidth}
         \centering
         \includegraphics[width=\linewidth]{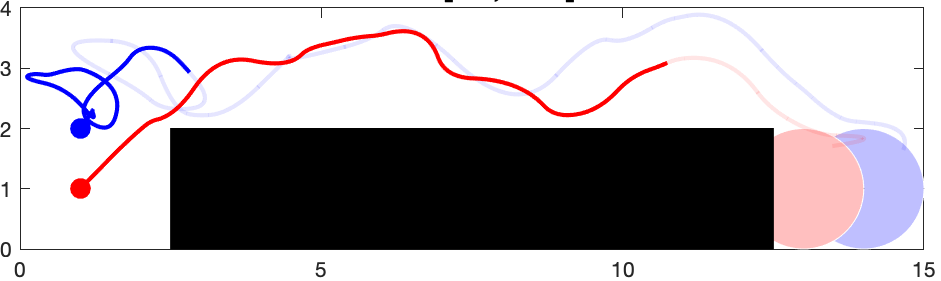}
         \caption{$\dt_1 = [0, 14.5]$s}
         \label{fig:ex4Seg1}
     \end{subfigure}
     \begin{subfigure}[b]{0.49\linewidth}
         \centering
         \includegraphics[width=\linewidth]{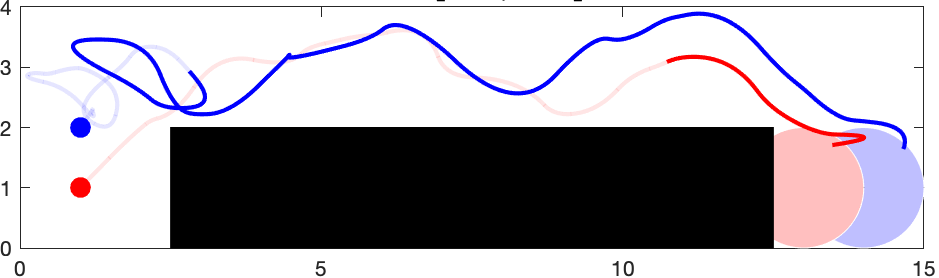}
         \caption{$\dt_2 = [14.5, 42.3]$s}
         \label{fig:ex4Seg2}
     \end{subfigure}
        \caption{Congested space: MAPS-RRT with 3 and 2 segments.}
        \label{fig:ex4Linear}
        \vspace{-1mm}
\end{figure}
\begin{figure}[b]
     \centering
     \begin{subfigure}[b]{0.32\linewidth}
         \centering
         \includegraphics[width=\linewidth]{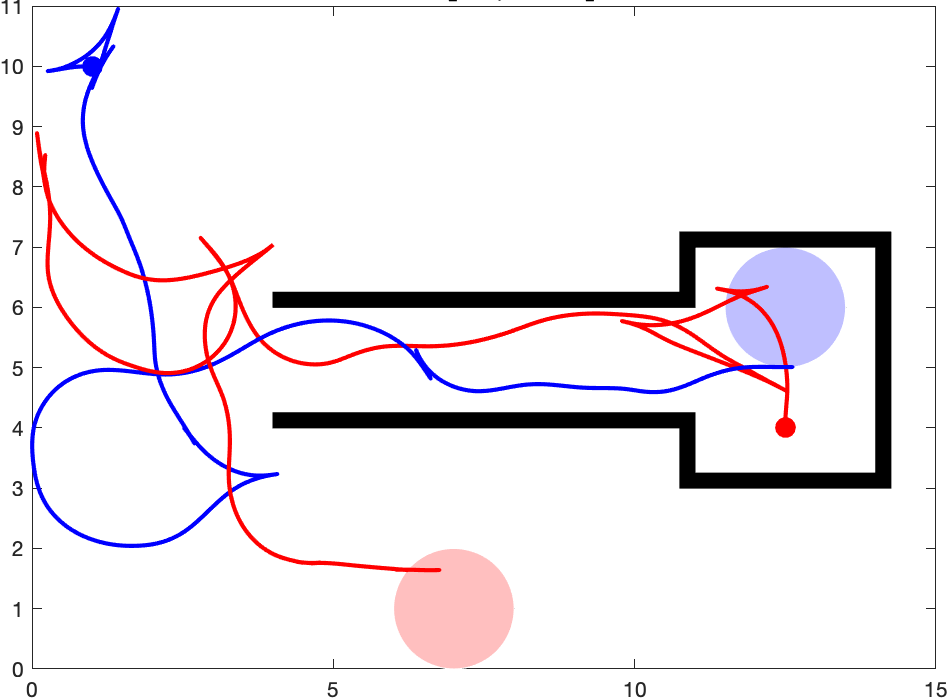}
        \caption{Full solution}
        \label{fig:ex3_2ndCar}
     \end{subfigure}
     \begin{subfigure}[b]{0.32\linewidth}
         \centering
         \includegraphics[width=\linewidth]{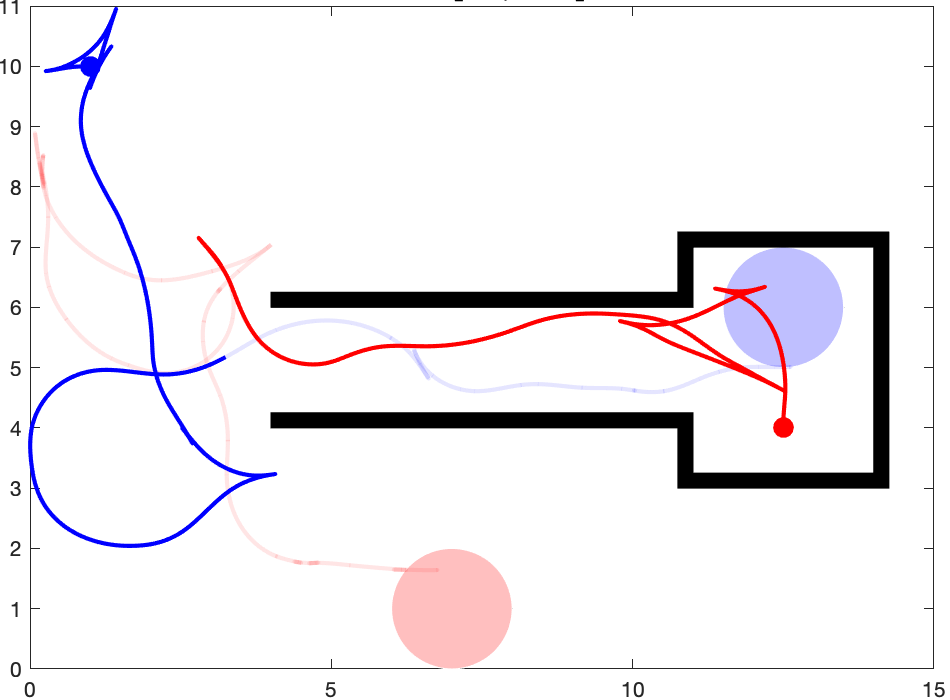}
         \caption{$[0, 24.5]$s}
         \label{fig:ex3Seg1}
     \end{subfigure}
     \begin{subfigure}[b]{0.32\linewidth}
         \centering
         \includegraphics[width=\linewidth]{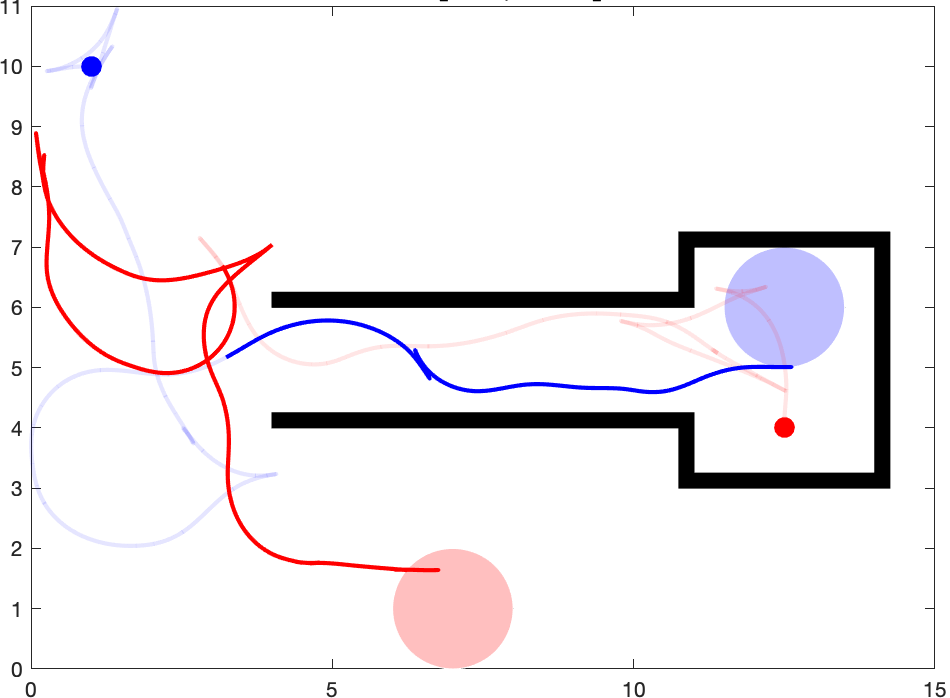}
         \caption{$[24.5, 41.2]$s}
         \label{fig:ex3Seg2}
     \end{subfigure}
     \caption{Corridor space: solution via MAPS-EST with $r=2$.}
     \label{fig:ex3_2ndCar_explained}
     \vspace{-1mm}
\end{figure}

In contrast, the environment in Fig. \ref{fig:ex2Unicylce} can easily admit non-intersecting paths as the blue agent can go around the red agent (Fig. \ref{fig:ex2_OneSegment}).  However, with RRT, we often get intersecting paths 
as shown in Fig. \ref{fig:ex2Seg1}-\ref{fig:ex2Seg2}.  By using MAPS-RRT and setting $r = 1$, we quickly get the easily-explainable solution in Fig. \ref{fig:ex2_OneSegment} with only one segment.
\begin{figure*}
     \centering
     \begin{subfigure}[b]{0.185\linewidth}
        \includegraphics[width=\linewidth]{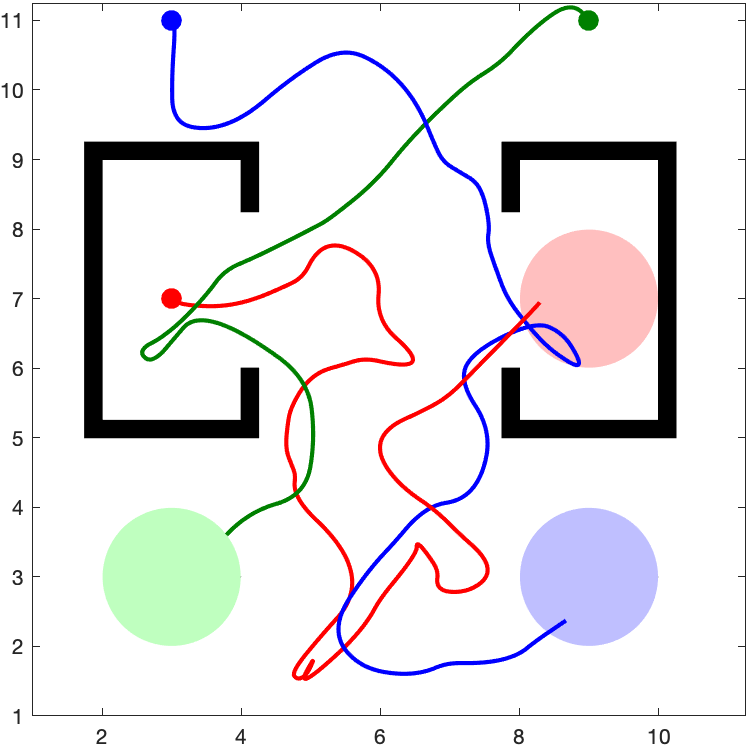}
        \caption{Full solution}
        \label{3agents_2ndOrderLinear}
     \end{subfigure}
     ~
     \begin{subfigure}[b]{0.185\linewidth}
         \centering
         \includegraphics[width=\linewidth]{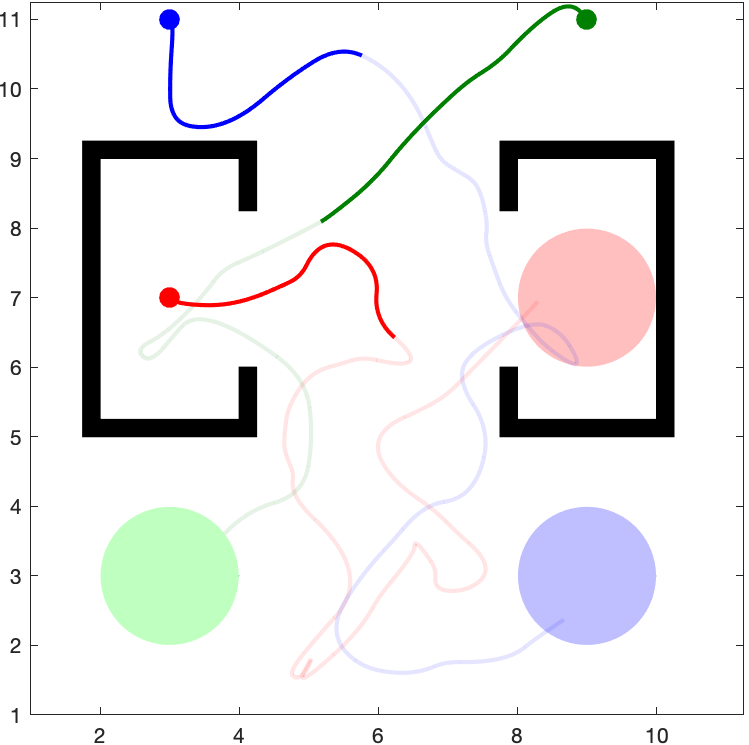}
         \caption{$\dt_1 = [0, 10.6]$s}
         \label{fig:3agentsSeg1}
     \end{subfigure}
     ~
     \begin{subfigure}[b]{0.185\linewidth}
         \centering
         \includegraphics[width=\linewidth]{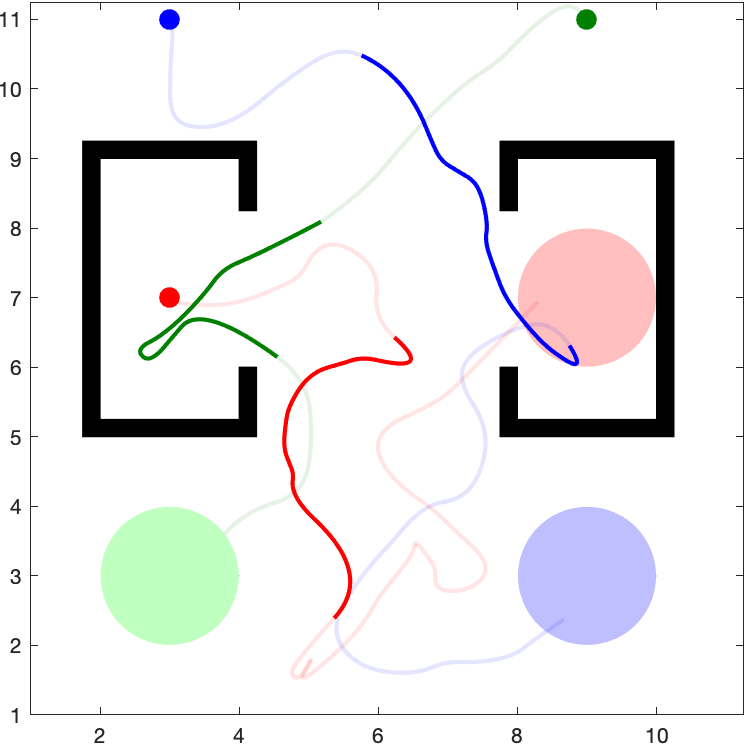}
         \caption{$\dt_2 = [10.6, 17.7]$s}
         \label{fig:3agentsSeg2}
     \end{subfigure}
     ~
     \begin{subfigure}[b]{0.185\linewidth}
         \centering
         \includegraphics[width=\linewidth]{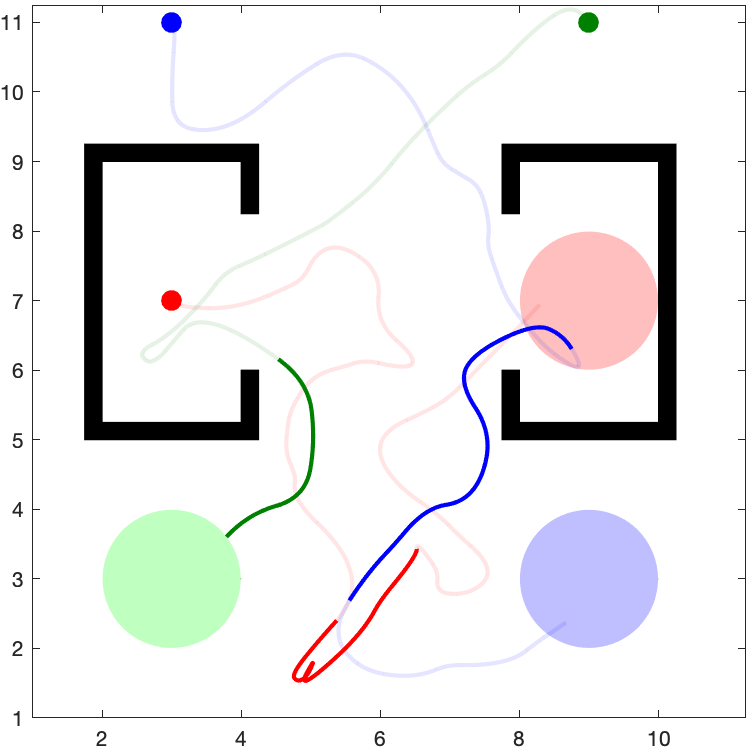}
         \caption{$\dt_3 = [17.7, 27.9]$s}
         \label{fig:3agentsSeg3}
     \end{subfigure}
     ~
     \begin{subfigure}[b]{0.185\linewidth}
         \centering
         \includegraphics[width=\linewidth]{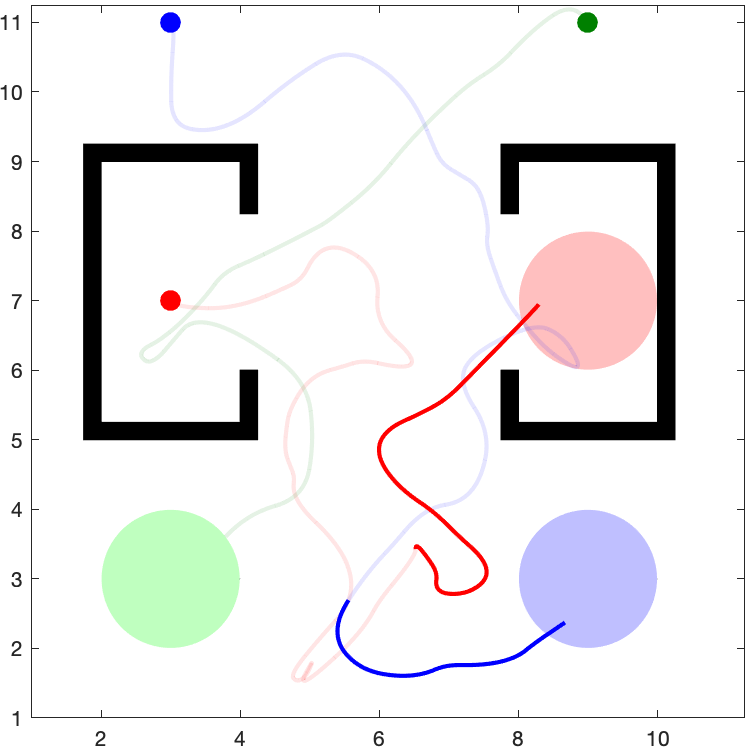}
         \caption{$\dt_4 = [27.9, 43.4]$s}
         \label{fig:3agentsSeg}
     \end{subfigure}
        \caption{Hallway space: solution via MAPS-RRT with 3 agents}
        \label{fig:3agents_explained_Linear}
        \vspace{-1mm}
\end{figure*}
\begin{table*}[t]
  \centering
  \resizebox{\textwidth}{!}{%
  \begin{tabu}{|c| l l c c c c c c c c c |}
    \hline
       & Space & Planner & Dyn. & \shortstack{\#\\ Agents} & r & \shortstack{ \# Solns. \\ Found (\%)} & \shortstack{Runtime \\ (s)} & \shortstack{Runtime \\ Success (s)} & \shortstack{Ave. \\ Cost} & \shortstack{Segment \\ Time  (s)} & $\shortstack{States added \\ per sec.}$ \\ [0.5ex] 
    \tabucline[1pt]{-}
    1& Open & MAPS-RRT & 1C & 2 & $\infty$ & 100 & $4 \pm 3$ & $4 \pm 3$ & $2 \pm 1$ & $0.8 \pm 0.5$ & $1300 \pm 75$ \\
    \hline
    2& Open & Lazy MAPS-RRT & 1C & 2 & $\infty$ & 100 & $3 \pm 2$ & $3 \pm 2$ & $2 \pm 1$ & $0.12 \pm 0.01$ & $1600 \pm 94$ \\
    \hline
    3& Open & MAPS-RRT & 1C & 2 & 1 & 90.5 & $11 \pm 9$ & $9 \pm 6$ & $1 \pm 0$ & $4 \pm 3$ & $800 \pm 99$ \\
    \hline
    4& Open & Lazy MAPS-RRT & 1C & 2 & 1 & 7.2 & $20 \pm 11$ & $12 \pm 9$ & $1 \pm 0$ & $2 \pm 1$ & $1200 \pm 77$ \\
    \tabucline[1pt]{-}
    5& Congested & MAPS-RRT & 1C & 2 & 1 & 85 & $37.9 \pm 33.1$ & $26.9 \pm 25$ & $1 \pm 0$ & $10.1 \pm 8.1$ & $629.4 \pm 135$ \\
    \hline
    6& Congested & Lazy MAPS-RRT & 1C & 2 & 1 & 0.0 & $100 \pm 0$ & n/a & n/a & $14.7 \pm 7.5$ & $598 \pm 42$ \\
    \tabucline[1pt]{-}
    7&Congested&MAPS-RRT&2C&2&$\infty$&100&$0.33\pm0.28$&$0.33\pm0.28$&$3\pm1$&$0.09\pm0.08$& $2100\pm152$ \\
    \hline
    8&Congested&MAPS-RRT&2C&2&3&100&$0.39\pm0.39$&$0.39\pm0.39$&$2\pm1$&$0.1\pm0.1$& $2000\pm182$ \\
    \hline
    9&Congested&MAPS-RRT&2C&2&1&100&$0.67\pm0.9$&$0.67\pm0.9$&$1\pm0$&$0.25\pm0.36$& $1400\pm262$ \\
    \tabucline[1pt]{-}
    10&Corridor&MAPS-RRT&1U&2&$\infty$&100&$0.53\pm0.58$&$0.53\pm0.58$&$3\pm1$&$0.13\pm0.15$& $1600\pm120$ \\
    \hline
    11&Corridor&MAPS-RRT&1U&2&4&100&$0.54\pm0.47$&$0.54\pm0.47$&$3\pm1$&$0.13\pm0.12$& $1600\pm110$ \\
    \hline
    12&Corridor&MAPS-RRT&1U&2&2&100&$0.77\pm0.84$&$0.77\pm0.84$&$2\pm0$&$0.21\pm0.26$& $1400\pm187$ \\
    \hline
    13&Corridor&MAPS-RRT&1U&2&1&100&$1.54\pm1.9$&$1.54\pm1.9$&$1\pm0$&$0.56\pm0.74$& $1000\pm182$ \\
    \tabucline[1pt]{-}
    14&Hallway&MAPS-RRT&2L&3&$\infty$&100&$4.15\pm3.9$&$4.15\pm3.9$&$3\pm1$&$1.13\pm0.9$& $1000\pm123$ \\
    \hline
    15&Hallway&MAPS-RRT&2L&3&3&100&$4.4\pm5.2$&$4.4\pm5.2$&$3\pm0$&$1.17\pm1.11$& $900\pm154$ \\
    \hline
    16&Hallway&MAPS-RRT&2L&3&2&100&$6.21\pm7.18$&$6.21\pm7.18$&$2\pm0$&$1.78\pm1.85$& $744\pm167$ \\
    \hline
    17&Hallway&MAPS-RRT&2L&3&1&93.5&$28.61\pm27.58$&$23.64\pm20.79$&$1\pm0$&$10.6\pm9.1$& $283\pm69$ \\
    \tabucline[1pt]{-}
  \end{tabu}}
  \caption{Benchmark results with 200 runs. The two character dynamics are $1^{st}$- and $2^{nd}$- order Linear, Unicycle, and Car, respectively. The maximum allowed time was $30s$ for rows 1-4, and $100s$ for rows 5-17. The mean and standard deviation over all planning runs are shown for columns 7-11.}
  \label{tab:table_benchmark_unlimited}
  \vspace{-2mm}
\end{table*}
Explainability can come at a cost of plan length, as we show 
in Fig.~\ref{fig:ex4Linear}: when planning with a bound of $r=3$ segments, we obtain the short plan in Fig.~\ref{fig:ex4Full3}, where the agents follow one another closely in the corridor. This example creates a zig-zag behavior between agents that follow $2^{nd}$-order linear dynamics, making it difficult for human validation (and indeed requires a 3-segment explanation).
However, a better-explainable plan, obtained by setting $r=2$, is given in~\ref{ex4_2ndOrderLinear}. There, the blue agent waits for the red agent to get far ahead, before entering the corridor, as is explained in Figs.~\ref{fig:ex4Seg1} and \ref{fig:ex4Seg2}.



When agents go in opposite directions through a corridor, as demonstrated in the environment of Fig. \ref{fig:ex3_2ndCar_explained}, a controller may want to assure that the planner does not get both robots in the corridor at the same time, as that would be risky (cause path crossing). The explanation given in Figs.~\ref{fig:ex3Seg1},~\ref{fig:ex3Seg2} shows that indeed, only one robot is at the corridor at a time.

Validating trajectories without explanations gets more difficult as the number of agents increases. For example, consider the solution shown in Fig.~\ref{3agents_2ndOrderLinear} where the agents 
have $2^{nd}$-order linear dynamics.
A human user could have a difficult time checking that the entire trajectory is collision free. However, planning with the MAPS framework presents an easily verified solution, as shown in Fig.~\ref{fig:3agents_explained_Linear}. Moreover, by decreasing bound $r$, we improve the explainability and as a result get a ``cleaner'' plan, e.g., the solution in Fig \ref{fig:explain-graph} was obtained with $r=3$. 

\subsection{Performance Evaluation}
We now turn to study the performance of our algorithms. Our results are summarized in Table~\ref{tab:table_benchmark_unlimited}. 
Recall that MAPS-X is paramterized by Planner \plannerX and inherits its planning properties. As this is an orthogonal concern to planning (for the scope of this paper), we focus on RRT as the planner. 

	
	

Our results confirm expected phenomena of explanations.
When planning without a bound ($r=\infty$), MAPS-RRT and Lazy MAPS-RRT perform the same search, but MAPS-RRT has additional computational cost of tracking segmentation. 
Thus, 
in e.g., Rows 1, 2 of Table~\ref{tab:table_benchmark_unlimited} 
both algorithms obtain the same average number of segments ($2\pm 1$ segments), but MAPS-RRT 
takes longer to compute the segmentations.
As the bound $r$ decreases,  
(Rows 3, 4, 8, 9, 11-13, and 15-17), 
MAPS-RRT takes longer to find solutions, but gives a lower number of segments.
For low values of $r$, Lazy MAPS-RRT is required to prune very often. Since unlike MAPS-RRT, it does not 
store segmentation information in the nodes, 
these prunings are expensive, resulting in the higher runtimes and fewer solutions found,
as can be seen in Rows 3-6.

\section{Conclusion and Discussion}
  \label{sec:conclusion}
  This work outlines a new aspect of \mmp by considering explanation schemes for plans. 
The MAPS-X framework can be readily plugged on to existing (centralized) planners in order to provide explainable plans. As we showed, using MAPS-X to find short explanations sometimes generates longer plans, but can also make plans neater (at the cost of computational time).
In future work, we plan to add explanation generation to state-of-the-art techniques, and in particular to decentralized approaches, in order to improve scalability. 

\clearpage{}
\bibliographystyle{IEEEtran}
\bibliography{references}

\end{document}